# Engineering of Hallucination in Generative AI: It's not a Bug, it's a Feature

TIM FINGSCHEIDT, PATRICK BLUMENBERG, BJÖRN MÖLLER
Institute for Communications Systems,
TU Braunschweig,
Schleinitzstraße 22, 38106 Braunschweig

Generative artificial intelligence (AI) is conquering our lives at lightning speed. Large language models such as ChatGPT answer our questions or write texts for us, large computer vision models such as GAIA-1 generate videos on the basis of text descriptions or continue prompted videos. These neural network models are trained using large amounts of text or video data, strictly according to the real data employed in training. However, there is a surprising observation: When we use these models, they only function satisfactorily when they are allowed a certain degree of fantasy (hallucination). While hallucination usually has a negative connotation in generative AI – after all, ChatGPT is expected to give a fact-based answer! – this article recapitulates some simple means of probability engineering that can be used to encourage generative AI to hallucinate to a limited extent and thus lead to the desired results. We have to ask ourselves: Is hallucination in generative AI probably not a bug, but rather a feature?

## 1  Introduction

It is all about *expectations*. Which expectations do we have towards a generative AI application? Widely accepted principles for the design of generative AI applications are helpfulness, honesty, and harmlessness (see, e.g., [1]). The opposite would be terms like "not helpful", untruthful, and toxic. Of course, there are many cases where users would like to obtain truthful information, e.g., in an information system application such as Google search. In other cases, however, fantasy is expected, if the user requests to generate a nice story about some prompted topics. But what does that mean? Fantasy literature has many subgenres. How far do we want to go? Are flying horses allowed? Or dragons? Or even new languages such as Elvish [2, Appendix E-F] or Klingon [3]? Even among English or German texts, there are many styles existing, just as many as there are authors. The same holds for videos and movies.

In summary, we can state that generative AI is expected to be *useful* for the user's purpose. It shall fulfill expectations. Accordingly, it is not always about helpfulness, honesty, and harmlessness. In the following, we shed light on two generative applications: In Section 2 we take a look under the surface of a large language model (LLM) application, and in Section 4 we do the same for a video generation AI such as GAIA-1 [4]. Section 3, as a sidestep, will provide some evidence that autoregressive models are operating in "dictation mode", as compared to a "writing mode", as they listen to what they are saying.

## 2  Large Language Models

A broader introduction into large language models (LLMs), their neural network topology, and their training processes can be found in [5, Chs. 7-9]. A brief and for our purposes sufficient introduction is provided by Fingscheidt et al. [6]. Here, we only recapitulate some of the LLM functional aspects that have significant influence on user experience and fantasy/hallucinations. These are (a) the autoregressive nature of an LLM, and (b) its probabilistic next token prediction. Figure 1 displays the autoregressive nature of an LLM, that consists of an attention-based transformer decoder [7, 8]. For simplicity, let's assume that the LLM input shown on the bottom of the figure as orange input buffer consists of *letter* tokens $y_{l-1}$, more specific, an entire query to the LLM may consist of the input token sequence $y_{1:l-1}$ with a number of *l-1* tokens. The response of the LLM to such query is then simply the autoregressive continuation of the query text, token-by-token. At first, the discrete probability distribution $\boldsymbol{P}_l$ for the next token with index $l$ is generated. It has as many probability elements $P_{l,\delta}$ as there are tokens in the token alphabet. The subsequent argmax function may now select the most probable next token $y_l$ that is then the next output token, but also appended to the query at the bottom of the figure when producing the follow-up token probability distribution $\boldsymbol{P}_{l+1}$, and so forth. By multiple autoregressive calls of the LLM, the system output sentence (response) is generated, until a special "end-of-sentence" token is being produced.

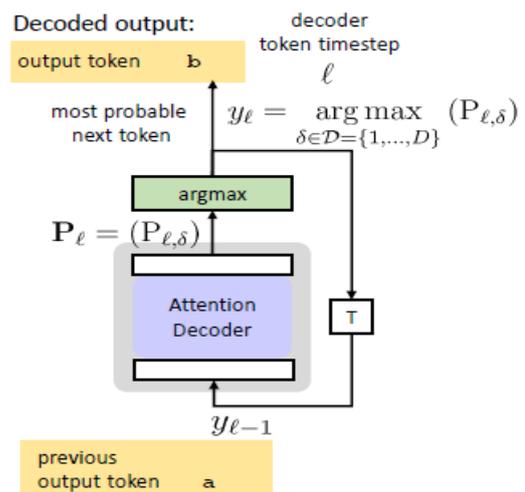

Figure 1: *Autoregressive nature of an attention-based large language model (LLM).*

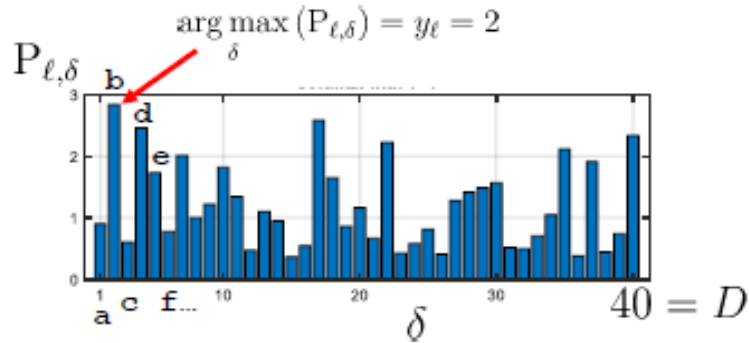

Figure 2: Next-token probability distribution, example.

Fig. 2 displays an example next-token probability distribution $\boldsymbol{P}_l = (P_{l,\delta})$. For simplicity, instead of the common byte-pair encodings (BPEs) [6, 9], we assume the token alphabet to only contain isolated letters, digits and a few symbols, in total a size of $D = 40$. In the example, the argmax operator selects the most probable token (letter) "b", which is 2[nd] in the token alphabet.

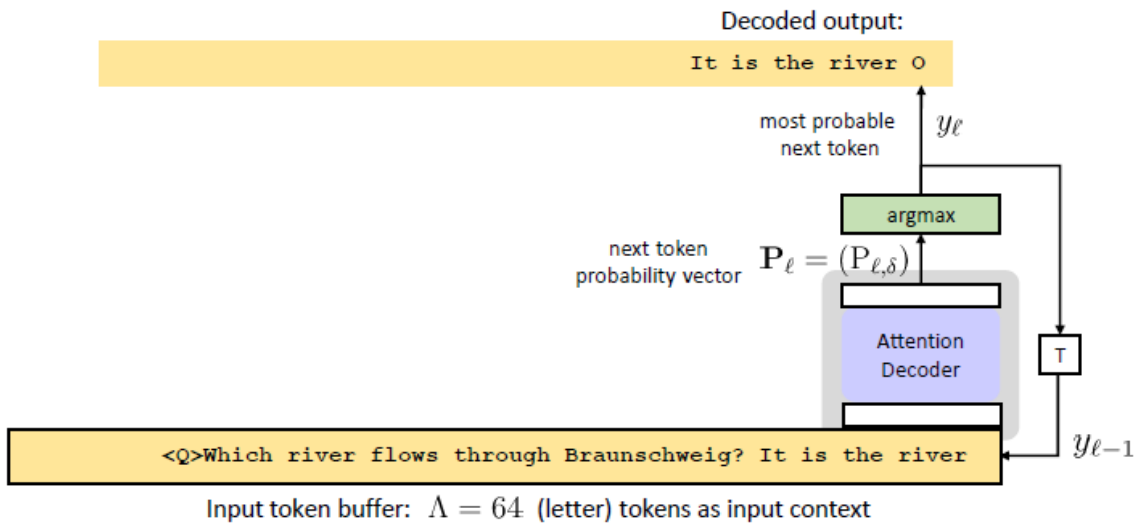

Figure 3: Example of a particular query and start of an answer.

Fig. 3 provides an example of a query "Which river flows through Braunschweig?" and the next 18 predicted letter tokens (including blanks) " It is the river O". Appending the just recognized letter "O" to the displayed input token buffer, the subsequently predicted token should be "k", as the river Oker flows through Braunschweig. Note that the input token buffer is of limited length. While in the example it is displayed with length $\Lambda = 64$, in GPT4 it already reaches $\Lambda = 32768$. So far, we provided a simplified sketch of the token-by-token autoregressive call structure in an attention decoder-based large language model.

In *training* of such a large language model, interestingly, the argmax operator is not even used. Instead, the principle of a generated pre-trained transformer (GPT) simply follows a distribution matching using the minimum cross entropy (MCE) as it is common in virtually all classification task trainings. MCE provides a Kullback-Leibler divergence between a one-hot target vector $\bar{P}_l$ (with all-zero elements $\bar{P}_{l,\delta} = 0$, except the one with the correct next token in the training text being $\bar{P}_{l,\delta^*} = 1$) and the predicted one $P_l$.

In *test* (also called inference or operation) of such a large language model, one could use an argmax operator to decide for the next token, just as we did in our simplified Figs. 1-3. Assuming the estimated probability distribution $P_l$ to be a good estimate of token posterior probabilities, according to Bayesian decision theory [10], the argmax delivers the next token *with lowest probability of error*. This is important to note, as in practice, LLMs *do not employ the argmax* operator but perform a couple of manipulations on $P_l$ and thereby deviate from the minimum error criterion in their next token prediction. *One could state that LLMs are parameterized to increase usefulness instead of truthfulness* (the latter being related to minimum error). These subsequently discussed manipulations govern the degree of hallucination.

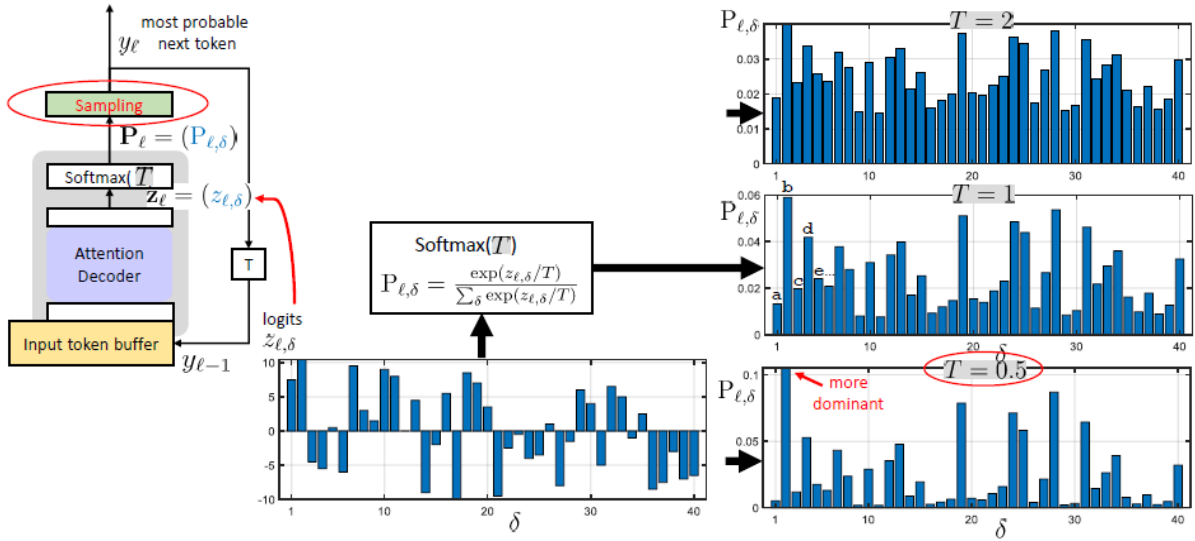

Figure 4: *Softmax in an LLM with illustration of the effect of temperature T.*

In particular, the LLM's processing steps are as follows: As detailed in Fig. 4, after the last attention decoder block, the so-called logits $z_{l,\delta}$ are subject to a *softmax* activation function (with temperature *T*), yielding the individual token probabilities according to

$$P_{\ell,\delta} = \frac{\exp(z_{\ell,\delta}/T)}{\sum_\delta \exp(z_{\ell,\delta}/T)}.$$

Temperature *T* is a hyperparameter that can be freely chosen in inference (in training it is commonly *T=1*), with the effects shown on the right-hand-side of Fig. 4: If the temperature is getting larger, the probability distribution $P_l = (P_{l,\delta})$ adopts a more uniform shape. Such a higher entropy makes the LLM output coming along with more fantasy, potentially even with newly created words. If the temperature is getting smaller, only the dominant peaks in the distribution survive, meaning that the output tokens are effectively drawn from only a small subset of tokens. In the case of $T \to 0$, only a one-hot vector $\mathbf{P}_l$ is predicted.

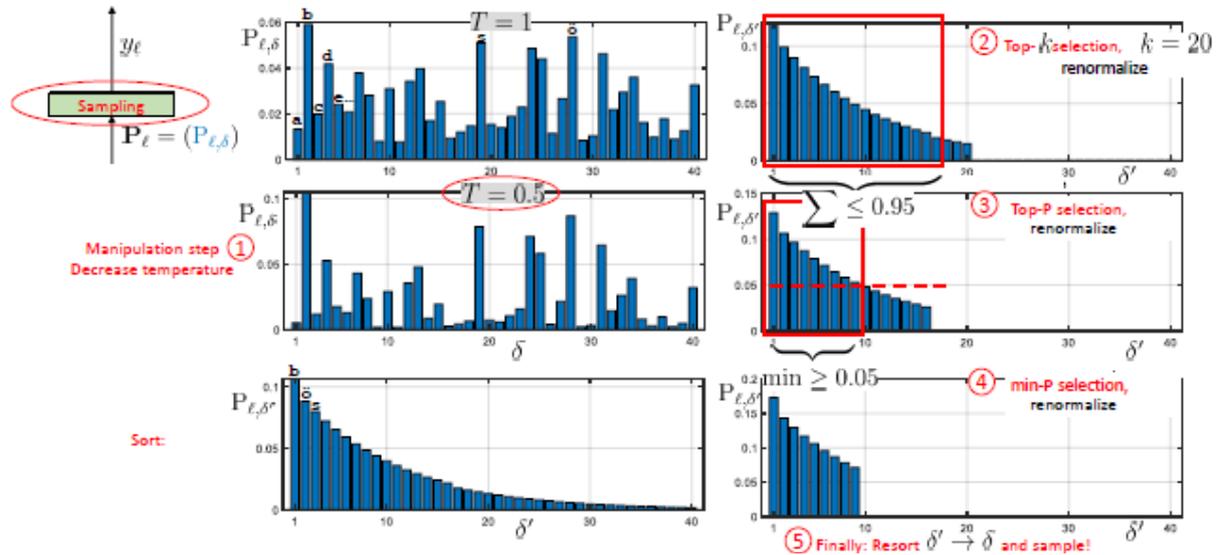

*Figure 5: Steps contained in the sampling procedure.*

As depicted in Fig. 4, instead of an argmax operation, the softmax operation is followed by a *sampling* procedure in LLM inference. Fig. 5 shows some internal details of the sampling procedure: (1) After the already discussed application of some temperature within the preceding softmax, probabilities are sorted from high to low (lower left) and (2) only the *top-k* (in the example: *k = 20*) probabilities survive. Among these, after renormalization to fulfill the stochastic constraints, (3) only the *top-P token probability mass* with *P = 95%* survives. After renormalization, (4) *min-P selection* is performed with *P = 5%*, allowing only tokens with minimum probability of *P* being selected. Finally, employing the remaining (shortened) probability distribution (in the example of length 9) in the lower right of Fig. 5, (5) *random sampling* from the distribution is performed—after having resorted the probabilities to their original token indices $\delta$. Beyond temperature *T*, we observe that the top-*k*, top-*P*, and min-*P* procedure along with the actual sampling allows for manipulation and shortening of the probability distribution so that sampling the next token in most of the cases leads to *useful* LLM output. We summarize: No minimum error is targeted, but sampling from a variety of preselected and therefore hopefully useful next tokens is how the inference is practically implemented. LLM outputs are outputs selected from random number generators!

Let us highlight the effect of an important hyperparameter, min-P. Employing a 4-bit-quantized, instruction-tuned Llama 3 LLM of size 3 B parameters [11], we choose T=0.8, Top-40, Top-0.95, and Min-0 (in short (0.8, 40, 0.95, 0) and provide the query "Which river flows through Braunschweig, Germany?" (ending with a blanc). The LLM response is "The river that flows through Braunschweig, Germany is the Ocker". Note the mistake in writing in the word "Ocker" (no river name), which should correctly be "Oker". A choice of Min-0 randomly draws from quite many tokens, so let's increase Min-P to Min-0.06, asking for an at least 6% probability of the next token. The LLM response stays the same, but delivers the response "Oste river" – which is a truly existing river in the region but still the wrong answer.

Being even more conservative by using Min-0.15 (i.e., only up to a handful of tokens may survive for sampling), the LLM's answer is: "The river that flows through Braunschweig, Germany is the Oder River, however, the Oder River does not flow through Braunschweig." First of all, as before, the Oder River is an existing river in Germany, but the wrong one. What makes this particular answer so interesting, is that during the autoregressive token-by-token production of the response, *the LLM starts reflecting its own output*, introduced by the word "however", followed by the exact opposite (now a correct) statement that the Oder River does not flow through Braunschweig. *This self-reflection can only be explained by the continuously and autoregressively appended LLM input buffer with previously sampled output tokens* – which appends the wrong answer to the query. The reaction of the LLM to its own wrong answer in this example is a negation of its own previous statement: *Although in this case, we do not observe a correction, but we see a mature, useful and also truthful reaction of the LLM to a wrong query (formerly, its own wrong answer).*

We summarize from this single example: *The choice of the softmax and sampling hyperparameters decides upon the degree of fantasy or truthfulness that we request from the LLM, but also influences the degree of the LLM's self-reflection.* Short resulting distributions increase truthfulness, since if the min-P criterion only allows a single token to pass, the sampling is equal to argmax, the latter, as previously discussed, minimizes the probability of error. In our example at hand, the LLM was not able to deliver the correct answer "Oker River", but when sampling from a short distribution, it negated its former wrong answer – a trend towards higher truthfulness.

## 3     Autoregressive Generative Models Listen to What they are Saying

*"The river that flows through Braunschweig, Germany, is the Oder River, however, the Oder River does not flow through Braunschweig."* In the previous section, we reported this LLM response to the query "Which river flows through Braunschweig, Germany?". This surprising LLM output is worth to be investigated a bit deeper. It is a sentence with a contradiction in itself, whereby the first part is obviously wrong (as the O*k*er River flows through Braunschweig,

Germany), whereas the second part of the sentence correctly contradicts the first one, however, without delivering the full truth (which would be the O*k*er river, that flows through Braunschweig).

There are not widely regarded verses in the New Testament, 1 Corinthians 1:14-16, where Apostle Paul wrote to the Corinthians that they shall not argue against each other to which spiritual leader or baptizer they belong. Here cited from the New International Version [12]," [14] I thank God that I did not baptize any of you except Crispus and Gaius, [15] so no one can say that you were baptized in my name. [16] Yes, I also baptized the household of Stephanas; beyond that, I don't remember if I baptized anyone else." Here, the parentheses in the NIV text were removed as they were anyway not in the underlying Greek source text. To calm down the discussions among Corinthian believers, in the first half of verse 14 (let's name it 14a), Paul claimed to have not baptized anybody in Corinth, while in the second half of this verse he quickly adds an exception, Crispus and Gaius. In the first part of verse 16, he adds another exception, which is the household of Stephanas. In the later part he provides a third statement that is either in contradiction to or at least it weakens the clear statement in verse 14a not to have baptized anybody in Corinth. In the end, he cannot tell the exact truth whom he actually baptized. The reader (or listener of the read-aloud epistle) will have noticed the sincerity and truthfulness of Apostle Paul in openly stepping back behind his original statement in verse 14a.

We observe a direct analogy to the response of the LLM as initially written in this section. What is the root cause of such a step-wise reflection of a wrong statement? We can formulate a hypothesis based on the analogy of our LLM response example with the cited verse from Paul as follows. It is well known since centuries about Paul's epistles that they have been dictated to a writer and have not been hand-written by himself. Exceptions give evidence to that, as we find one in 1 Cor. 16:21: "[21] I, Paul, write this greeting in my own hand." So, as we can assume that 1 Cor. 1:14-16 have been dictated by Paul, it is understandable that he reflects his own just dictated sentences as long as they reside in his auditory buffer. This is nicely summarized in the famous saying "How can I know what I think till I see what I say?" (supposedly originally from [13]). This is not only true for Apostle Paul, quickly dictating exceptions about his no-baptism statement in verse 14a, but also to our LLM, which quickly states that its own previous statement was wrong. In both cases, there was a feedback loop involved: on the one side Paul's auditory buffer (in case he dictated the entire verses 14-16), and on the other side the LLM's autoregressive feedback loop that allowed both to "see what [they] say".

In summary, we can hypothesize that vanilla-type decoder-based LLMs actually *don't write their output*, although most of their training material is from written text. *Instead, they dictate their output*, without correction capabilities (a word being said is said), but with self-reflection capabilities, allowing some flexibility of thought given a well-chosen set of sampling hyperparameters. These reflection capabilities are technically provided by their autoregressive operation that appends the system output to the query.

## 4      Generative Video Models

After having discussed various aspects of hallucinations, feedback, and self-reflection both in humans and in LLMs in the previous sections, here we would like to explore the influence of one of the previously introduced sampling hyperparameters in the context of generative video AI. One of the most prominent video generation systems for automotive driving scenes is GAIA-1 [4], in the meantime already existing in a follow-up version GAIA-2 [14]. As shown in each row of Figure 6, when being prompted by the leftmost image, GAIA-1 produces consistent videos under the given environment, light, and weather condition.

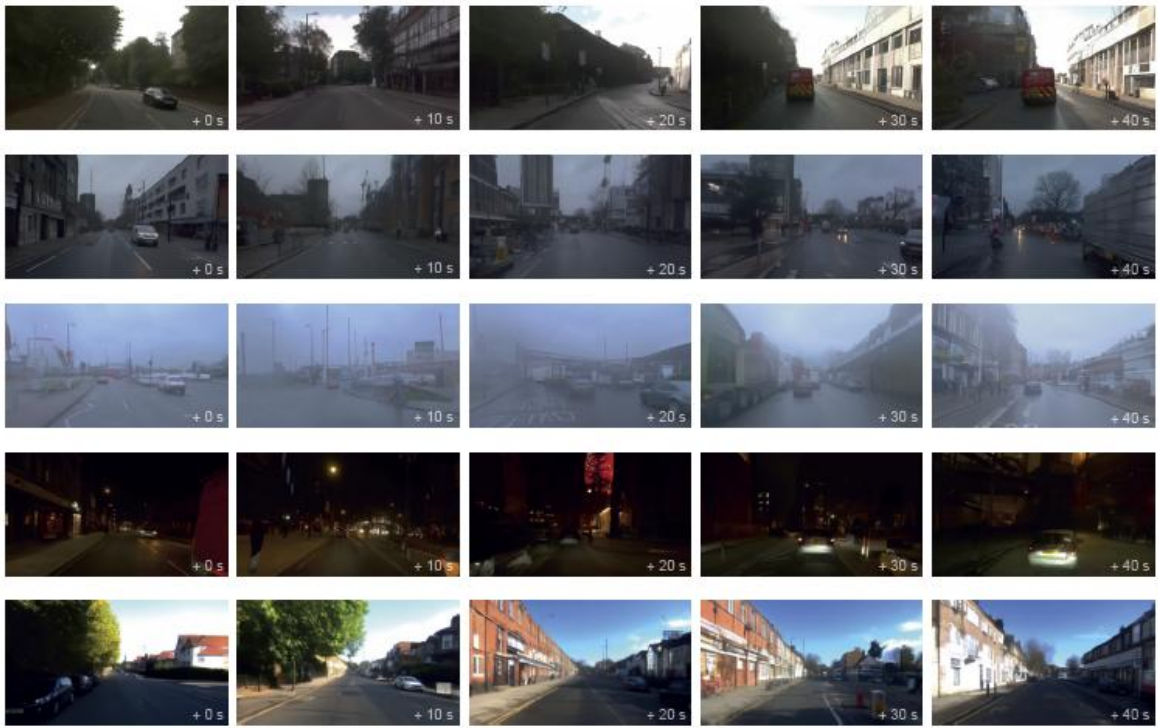

*Figure 6: Each row displays GAIA-1-generated frames sampled each 10s, taken from [4].*

Now let us obtain a brief overview, how GAIA-1 technically works.

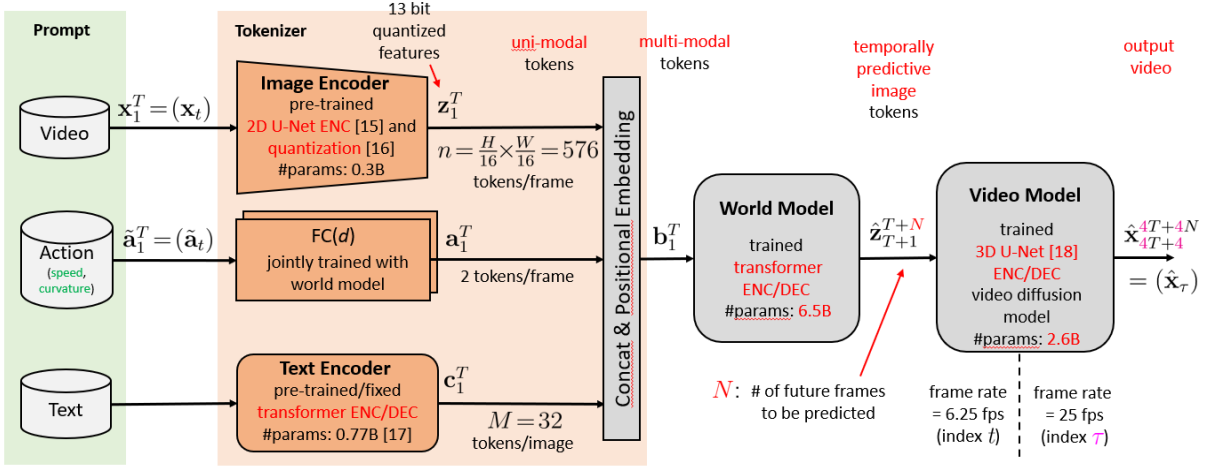

*Figure 7: Block diagram of the GAIA-1 video generation system.*

Figure 7 displays on the left-hand side ("prompt") the three optional input modalities of GAIA-1, which are images or videos $\boldsymbol{x}_1^T$, actions $\widetilde{\boldsymbol{a}}_1^T$, or text. All of these optional prompt modalities are subject to individual encoding by either a pre-trained 2D U-Net [15] with quantization [16] to obtain discrete output tokens $\boldsymbol{z}_1^T$ (representing images), or by a fully connected (FC) layer trained from scratch to represent speed and curvature (e.g., "left") action tokens, or by a transformer-based text encoder [17]. After concatenation and positional embedding, the multimodal discrete token sequence $\boldsymbol{b}_1^T$ is input to a transformer-based autoregressively called world model, that outputs a discrete image token sequence $\widehat{\boldsymbol{z}}_{T+1}^{T+N}$, representing $N$ predicted future images. Finally, in GAIA-1, these are input to a video diffusion model [18] that performs 4-times temporal oversampling from GAIA-1's native 6.25 frames per second (fps) operation to 25 fps at the output, along with temporally coherent output video generation $\widehat{\boldsymbol{x}}_{4T+4}^{4T+4N}$ from the input image token sequences.

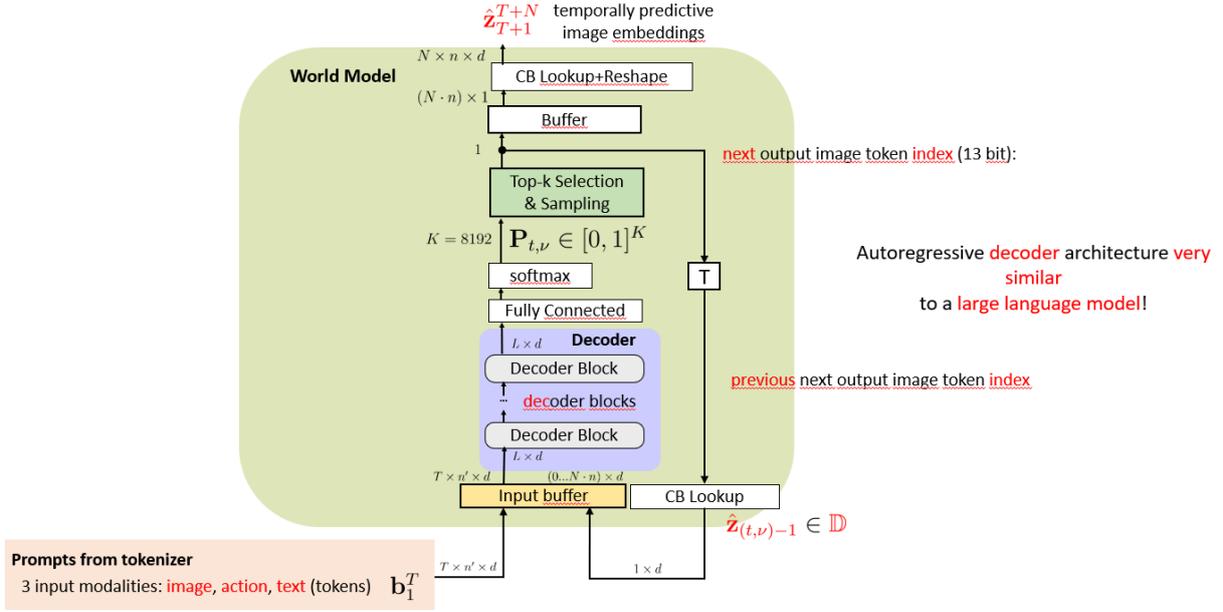

*Figure 8: Block diagram of the world model in GAIA-1.*

In Figure 8, we can take a deeper look into GAIA-1's Figure 7 world model. As with the LLM in Figure 1, there is an autoregressive call of an attention-based decoder model. Corresponding to the image encoder's tokens in Figure 7, also the tokens predicted here, refer to image patches (an image consists of 576 patches). As in Figure 4, we identify softmax (with internal temperature *T*) and a sampling procedure, the latter employing the typical probability distribution manipulations as they were depicted in Figure 5.

*From a technical viewpoint, we can summarize that while an LLM is a text-input text-output next text-token predictor, GAIA-1 is a multimodal-input video-output next image predictor. GAIA-1's world model is at the heart of it and directly corresponds to the LLM having the autoregressive call structure in common.*

In analogy to Section 2, now we are interested in exploring sampling hyperparameters. Here, in the video generation application, we would like to put our focus on the Top-*k* sampling procedure. As GAIA software code is not publicly available, we build our experiments on our own automotive video generation system that is inspired by GAIA-1. Our video-only input approach operates at 4 fps on quadratic 256 x 256 image videos, sharing with GAIA-1 the 13 bit image patch tokenization. For details, the reader is referred to [19].

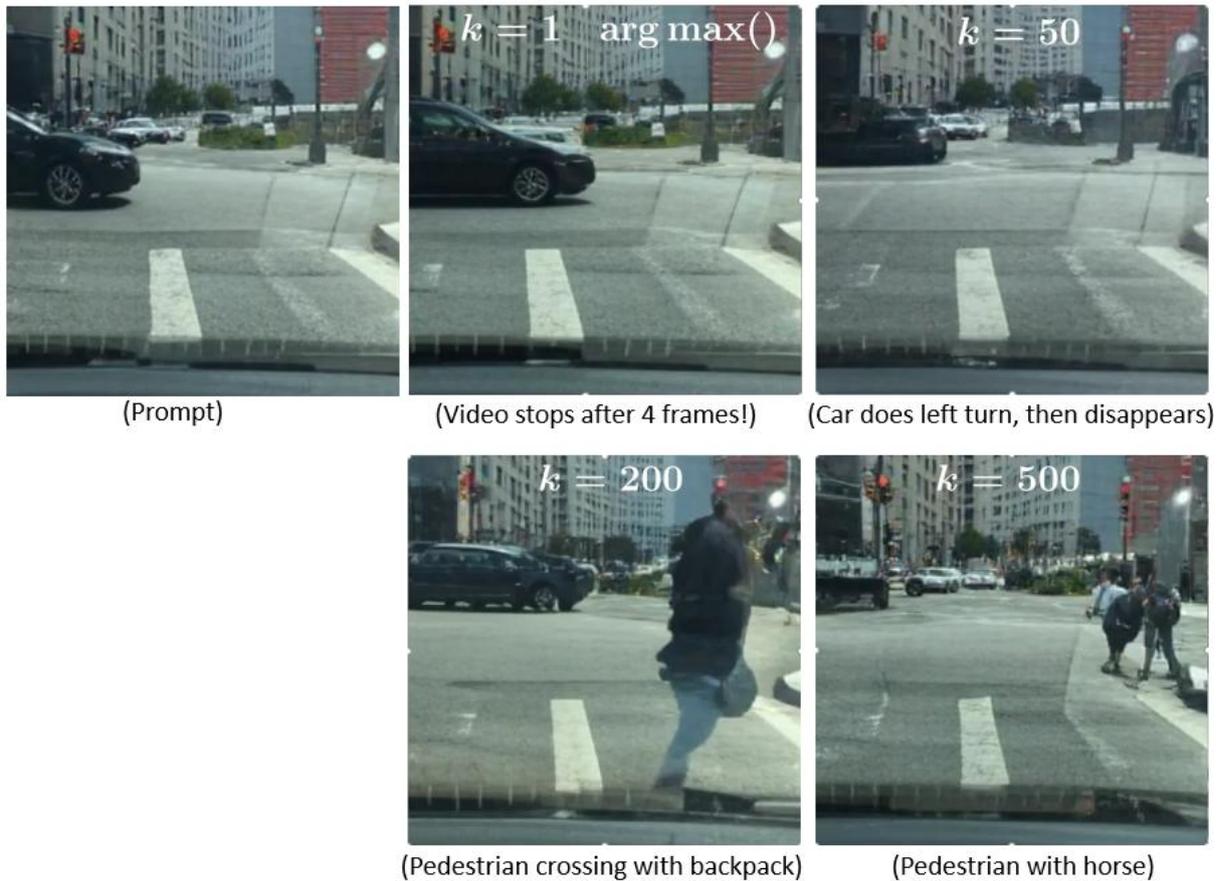

*Figure 9: Examples of Top-k influence in automotive video generation.*

Figure 9 depicts in the upper left an image prompt to our video generator (actually it is prompted by two consecutive frames), while the other four images are the predicted ones after 5.25s of video prediction at various values $k$ of the Top-$k$ sampling procedure. For $k = 1$ (equivalent to the use of argmax), the predicted video freezes after four frames. For $k = 50$ (the default GAIA-1 setting), the crossing car does a left turn and then disappears. For $k = 200$ representing already high hallucinating power, a pedestrian with backpack and an additional bag crosses the road. Finally, the choice of $k = 500$ underpins its fantasy-oriented generation by predicting a pedestrian with a horse crossing the street on the right-hand side, and then mounting on the horse right in the center of the crossing.

But why does $k = 1$ lead to a freeze of the predicted video? Its equivalence to using the argmax means that this hyperparameter choice actually minimizes error in future frame prediction! This effect is known for a while in any minimum mean square error (MSE) estimation that relies on temporal extrapolation (like our generative methods do), as the fewer information is available, MSE estimation tends towards delivering the (conditional) mean: As described by Fingscheidt et al. [20], this effect has been investigated for speech transmission as *muting effect* (zero is the mean of audio samples), while here we observe that on pixel level after four frames (i.e., after 1s) the actual available information is so low, that the conditional mean of each pixel value is generated – conditioned on the previous frame's pixel value, which is simply the same pixel value as in the image before. This results in freezing of the predicted video.

For *k* = 1, the system is asked to operate under a minimum error requirement. As we know from employees who have to work under a zero-error expectation, they may simple become blocked (as the frozen video): As an example, Kriegesmann et al. [21], write: "Innovative management and risk friendliness are necessary, but the way in which failure is handled and the resulting fear of making mistakes block the (innovative) efforts of specialists and managers." *In summary, both humans and generative AI approaches operate in a most useful manner under a non-zero error target, which means some form of random sampling, hallucination, or simply fantasy.*

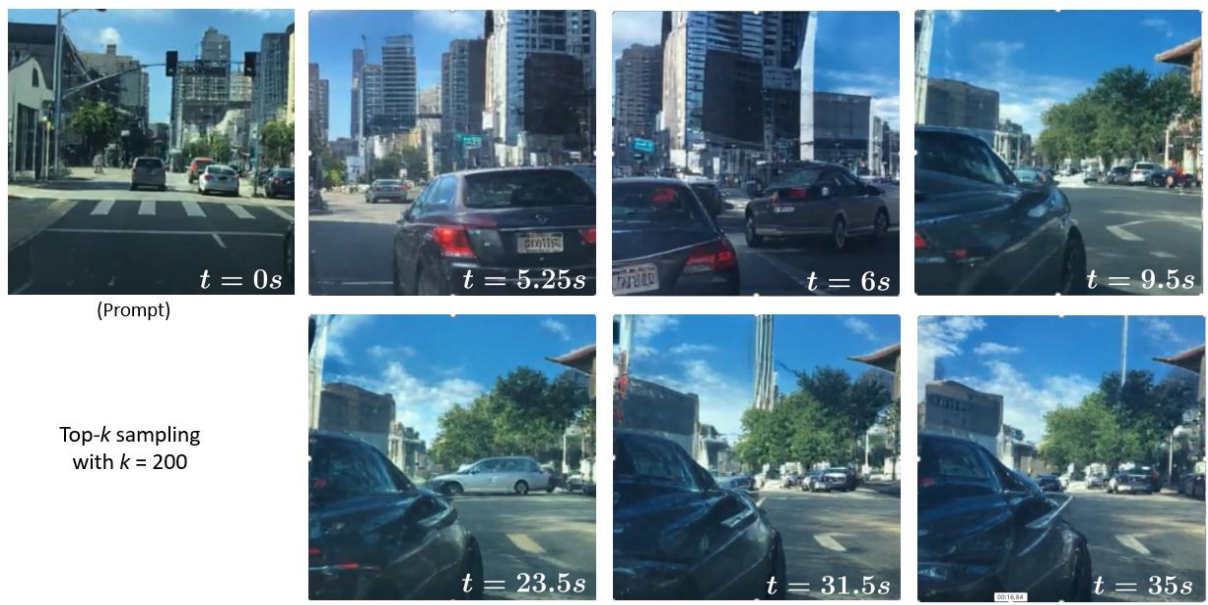

*Figure 10: Sampled frames of an example video for Top-k sampling with k = 200.*

Now let's be more creative and choose *k* = 200. Figure 10 shows image samples of a single predicted video with this Top-*k* sampling hyperparameter choice. Again, in the upper left, we see the last image of a two-image video prompt. At time *t* = 5.25s, a car passes by on the right-hand side, at *t* = 6s, the ego vehicle starts a right turn just behind the passing car, at *t* = 9.5s, the ego vehicle stops just behind (or crashes into) a car standing in front. At *t* = 23.5s, a light grey car passes by on the crossing – so although the ego vehicle is still not in motion, the video didn't freeze. At *t* = 31.5s, the scenery becomes supernatural: A thin skyscraper enters the horizon from the left side, which, at *t* = 35s turns to a pole and moves further to the right. For practical applications, this means that while generating corner cases such as a crash or any other traffic accident is highly useful, e.g., for validation of autonomous driving functions, the skyscraper and the pole at the horizon in this example are not drawn from the tails of any real data distribution, instead, they are simply impossible in reality. *In summary, there are Top-k sweet* *spots for a useful degree of hallucination, which in our video generation examples is around k = 50* [4]*, but can still be as high as k = 1000* [19]*.*

## 5   Concluding Remarks

This article has explored technical aspects of large language models (LLMs) and video generative AI (GAIA-1) with a focus on the question: How can we control the degree of hallucination? Is minimum-error a useful target for inference of generative AI? Indeed, it is not, thereby identifying hallucination to be a feature rather than a bug. In fact, there are hyperparameters that need to be tuned with care to optimize for practical usefulness of generative AI applications. By exemplary comparison of LLM behavior in certain settings to a dictated text from Apostle Paul in the New Testament, we hypothesize that *LLMs actually don't write their output text. Instead, they dictate it*, and at the same time listen to it, appending it into their query buffer and thereby allowing to self-reflect what they just predicted. In that sense, a certain amount of hallucination along with the just described autoregressive operation of decoder-based LLMs allows them to oppose their own earlier wrong statements and thereby improve their truthfulness.

Only recently, LLM reasoning started to become a vital area of research. Through post-training techniques based on reinforcement learning, an LLM can learn to find answers to complex questions through a step-by-step thought process [22]. The LLM reasons about the query in natural language and continuously reflects on its own output. These techniques can strengthen the observed self-reflection towards a self-correcting behaviour [23] and help LLMs overcome some of the limitations inherent in autoregressive generation.

## References

[1] A. Askell, Y. Bai, A. Chen et al., "A General Language Assistant as a Laboratory for Alignment," arXiv:2112.00861, Dec. 2021.
[2] J. R. R. Tolkien, The Lord of the Rings: The Return of the King. London: George Allen & Unwin, 1955.
[3] M. Okrand, The Klingon Dictionary. New York, NY, USA: Pocket Books, 1985.
[4] A. Hu, L. Russell, H. Yeo, Z. Murez, G. Fedoseev, A. Kendall, J. Shotton, and G. Corrado, "GAIA-1: A Generative World Model for Autonomous Driving," arXiv:2309.17080, Sep. 2023.
[5] D. Jurafsky and J. H. Martin, Speech and Language Processing: An Introduction to Natural Language Processing, Computational Linguistics, and Speech Recognition, with Language Models, 3rd ed. Online manuscript, Aug. 24, 2025. [Online]. Available: https://web.stanford.edu/~jurafsky/slp3/

[6] T. Fingscheidt, Z. Li, and T. Lohrenz, "From End-to-End Automatic Speech Recognition to ChatGPT: A Technical Journey," *Jahrbuch der Braunschweigischen Wissenschaftlichen Gesellschaft,* vol. 2023. Braunschweig, Germany: Braunschweigische Wissenschaftliche Gesellschaft, 2024, pp. 233–244.

[7] A. Vaswani, N. Shazeer, N. Parmar, J. Uszkoreit, L. Jones, A. N. Gomez, L. Kaiser, and I. Polosukhin, "Attention Is All You Need," in Proc. of NIPS, Long Beach, CA, USA, Dec. 2017, pp. 1–11.

[8] A. Radford, K. Narasimhan, T. Salimans, I. Sutskever et al., "Improving Language Understanding by Generative Pre-Training," Jun. 2018.

[9] R. Sennrich, B. Haddow, and A. Birch, "Neural Machine Translation of Rare Words with Subword Units," in Proc. of ACL, Berlin, Germany, Aug. 2016. pp. 1715–1725.

[10] R. O. Duda, P. E. Hart, and D. G. Stork, "Bayesian Decision Theory," in *Pattern Classification*, 2nd ed., New York, NY, USA: Wiley-Interscience, Oct. 2000, pp. 20–83.

[11] A. Grattafiori, A. Dubey, A. Jauhri et al., "The Llama 3 Herd of Models," arXiv:2407.21783, Jul. 2024.

[12] Holy Bible, New International Version, Fully Revised Edition, NIV Study Bible, Zondervan, 2011, 1 Cor. 1:14–16.

[13] G. Wallas, The Art of Thought. New York. NY, USA: Harcourt, Brace and Co., 1926.

[14] L. Russell, A. Hu, L. Bertoni, G. Fedoseev, J. Shotton, E. Arani, and G. Corrado, "GAIA-2: A Controllable Multi-View Generative World Model for Autonomous Driving," arXiv:2503.20523, Mar. 2025.

[15] O. Ronneberger, P. Fischer, and T. Brox, "U-Net: Convolutional Networks for Biomedical Image Segmentation," In Proc. of MICCAI, Munich, Germany, Oct. 2015, pp. 234-241.

[16] A. van den Oord, O. Vinyals, and K. Kavukcuoglu, "Neural Discrete Representation Learning," In Proc. of NeurIPS, Long Beach, CA, USA, Dec. 2017, pp. 6309-6318.

[17] C. Raffel, N. Shazeer, A. Roberts, K. Lee, S. Narang, M. Matena, Y. Zhou, W. Li, P. J. Liu, "Exploring the Limits of Transfer Learning with a Unified Text-to-Text Transformer," Journal of Machine Learning Research, vol. 21, no. 140, pp. 1-67, Jun. 2020.

[18] J. Ho, T. Salimans, A. Gritsenko, W. Chan, M. Norouzi, and D. J. Fleet, "Video Diffusion Models," In Proc. of NeurIPS, New Orleans, LA, USA, Nov. 2022, pp. 8633-8646.

[19] B. Möller, Z. Li, M. Stelzer, T. Graave, F. Bettels, M. Ataya, and T. Fingscheidt, "OpenViGA: Video Generation for Automotive Driving Scenes by Streamlining and Fine-Tuning Open Source Models with Public Data," arXiv:2509.15479, Sep. 2025.

[20] T. Fingscheidt and P. Vary, "Softbit Speech Decoding: A New Approach to Error Concealment," IEEE Transactions on Speech and Audio Processing, vol. 9, no. 3, pp. 240-251, Mar. 2001.

[21] B. Kriegesmann, T. Kley, and M. G. Schwering, "Making Organizational Learning Happen: The Value of "Creative Failures"," Business Strategy Series, vol. 8, no. 4, pp. 270-276, May 2007.

[22] D. Guo, D. Yang, H. Zhang, *et al.,* "DeepSeek-R1 Incentivizes Reasoning in LLMs Through Reinforcement Learning," *Nature*, vol. 645, pp. 633–638, Jul. 2025.

[23] R. Ma, P. Wang, C. Liu, et al., "S²R: Teaching LLMs to Self-Verify and Self-Correct via Reinforcement Learning," In Proc. of ACL, Vienna, Austria, Jul. 2025, pp. 22632–22654.